\documentclass{article}

\usepackage{microtype}
\usepackage{graphicx}
\usepackage{subcaption}
\usepackage{booktabs} %

\usepackage{hyperref}

\usepackage[preprint]{icml2026}

\usepackage{amsmath}
\usepackage{amssymb}
\usepackage{mathtools}
\usepackage{amsthm}

\usepackage[capitalize,noabbrev]{cleveref}

\theoremstyle{plain}

\theoremstyle{definition}

\theoremstyle{remark}

\newcommand{\mysubsub}[1]{\noindent\textbf{#1}:}

\usepackage[textsize=tiny]{todonotes}
\usepackage{multirow}

\icmltitlerunning{Submission and Formatting Instructions for ICML 2026}

\begin{document}

\twocolumn[
  \icmltitle{Say, Dream, and Act: 
  Learning Video World Models for \\ 
  Instruction-Driven Robot Manipulation}

  \icmlsetsymbol{equal}{*}

  \begin{icmlauthorlist}
    \icmlauthor{Songen Gu}{equal,yyy}
    \icmlauthor{Yunuo Cai}{equal,yyy}
    \icmlauthor{Tianyu Wang}{yyy}
    \icmlauthor{Simo Wu}{yyy}
    \icmlauthor{Yanwei Fu}{yyy}
  \end{icmlauthorlist}

  \icmlaffiliation{yyy}{Fudan University}

  \icmlkeywords{Machine Learning, ICML}

  \vskip 0.3in
]

\printAffiliationsAndNotice{\icmlEqualContribution}

\begin{abstract}
Robotic manipulation requires anticipating how the environment evolves in response to actions, yet most existing systems lack this predictive capability, often resulting in errors and inefficiency. While Vision-Language Models (VLMs) provide high-level guidance, they cannot explicitly forecast future states, and existing world models either predict only short horizons or produce spatially inconsistent frames. To address these challenges, we propose a framework for fast and predictive video-conditioned action. Our approach first selects and adapts a robust video generation model to ensure reliable future predictions, then applies adversarial distillation for fast, few-step video generation, and finally trains an action model that leverages both generated videos and real observations to correct spatial errors. Extensive experiments show that our method produces temporally coherent, spatially accurate video predictions that directly support precise manipulation, achieving significant improvements in embodiment consistency, spatial referring ability, and task completion over existing baselines. Codes\&Models will be released.

\end{abstract}

\section{Introduction}

Robotic manipulation in real-world environments is challenging because it requires both understanding the current scene and anticipating how it will evolve. Humans naturally predict object motion, spatial relationships, and environmental changes, allowing smooth and adaptive manipulation. Most robotic systems, however, lack this predictive ability, often resulting in errors, inefficiency, or unsafe behavior. Therefore, models that can forecast future states are essential for precise, long-horizon control.

Recent Vision-Language Models (VLMs) demonstrate strong decision-making capabilities~\cite{zitkovich2023rt,kim24openvla}, conditioning actions on high-level instructions or observations. Despite their promise, VLMs often fail at producing accurate low-level actions due to limited spatial reasoning and slow inference~\cite{black2024pi_0,gr00tn1_2025,wen2025dexvla}. Crucially, they do not explicitly model the consequences of actions, which limits their effectiveness in complex, multi-step manipulation tasks.

World models, which predict future observations~\cite{du2023learning,tian2024predictive,zhang2025generativevisualforesightmeets,li2025unified}, provide a more suitable solution. By forecasting how the environment evolves over time, they supply rich visual context that downstream action models can leverage for informed decision-making. Despite their promise, current world model approaches face several critical limitations. (1)\textit{Limited temporal horizon}: many models can only predict a short sequence of future frames, necessitating repeated rollouts to handle long-horizon tasks~\cite{hu2024video}. (2)\textit{Poor spatial consistency}: predictions often suffer from distortions or inconsistent robot embodiments, compromising spatial accuracy and reducing their usefulness for precise manipulation. (3)\textit{High computational cost}: while large video generation models achieve high predictive quality, their iterative nature incurs substantial computational cost, making real-time deployment challenging~\cite{jang2025dreamgen}.

On the other hand, diffusion-based video generators achieve strong predictive quality, but rely on iterative denoising, which is computationally expensive. Techniques such as distillation and adversarial training~\cite{salimans2022progressive,sauer2024adversarial,zhang2024sf} can accelerate inference, but maintaining spatial and temporal fidelity—crucial for manipulation—remains challenging.

To address these challenges, we propose  a Dream4manip  framework of \textbf{Say, Dream, and Act} that integrates high-fidelity video world models with an in-context conditioned action model for instruction-driven robot manipulation. Our approach is guided by three core principles:  
\emph{Say} of learning a robust video-based world model capturing task-relevant dynamics;  
\emph{Dream} of enabling length-agnostic imagination of future outcomes via frame-rate-agnostic video prediction; and  
\emph{Act} of producing executable actions by treating imagined trajectories as in-context examples grounded in real observations.

Formally, our method proceeds in three stages:  
(1) \textbf{World model selection and adaptation:} We benchmark several state-of-the-art video generation models on metrics such as embodiment consistency, spatial referring ability, and task completion, and select a strong foundation model. Domain adaptation ensures reliability across new scenes, camera viewpoints, and robot embodiments.  
(2) \textbf{Adversarial distillation for fast generation:} We distill the adapted model using latent-space adversarial loss and reconstruction supervision, enabling few-step denoising that preserves spatial and temporal fidelity while reducing computational cost.  
(3) \textbf{In-context conditioned action modeling:} We train an action model that consumes both imagined trajectories and real observations, correcting spatial errors in the world model and producing accurate action sequences.

Our approach is motivated by the need for robots to anticipate and reason about future states in complex, dynamic environments. By integrating high-fidelity video prediction with in-context action generation, we enable reliable and efficient manipulation in cluttered scenes. \textbf{Say, Dream, and Act} unifies world modeling and action prediction into a single, fast policy capable of imagining, anticipating, and acting robustly across diverse tasks. Extensive experiments demonstrate state-of-the-art performance in long-horizon manipulation, with improvements in Embodiment Consistency (EC), Referring Success Rate (RSR), Interaction Success Rate (ISR), and Task Completion Rate (TCR). In particular, Cosmos2 models, after domain adaptation and distillation, significantly outperform baselines in EC and RSR, highlighting the critical role of accurate and robust world models for instruction-driven robotic control.

\noindent\textbf{Contributions}. 
We make several contributions toward enabling fast, long-horizon robot manipulation. First, we develop a robust video-based world model by selecting and adapting a state-of-the-art video generation backbone (\textsc{Cosmos-Predict2}) to the robotic domain, applying domain adaptation and latent-space adversarial distillation to achieve high-fidelity predictions with few-step denoising. Second, we introduce a length-agnostic imagination mechanism that compresses arbitrary-length execution trajectories into a fixed set of keyframes, enabling frame-rate-agnostic prediction of long-horizon outcomes without relying on low-level action alignment. Third, we propose an in-context conditioned action model that treats imagined trajectories as examples while grounding predictions in real observations, allowing the system to correct spatial errors and produce executable actions. Finally, we integrate these components into a unified framework: \textbf{Say, Dream, and Act} that combines predictive world modeling and action reasoning into a single, fast policy capable of anticipating, planning, and executing manipulation tasks across diverse environments.

\begin{figure*}
  \centering
   \includegraphics[width=0.9\linewidth]{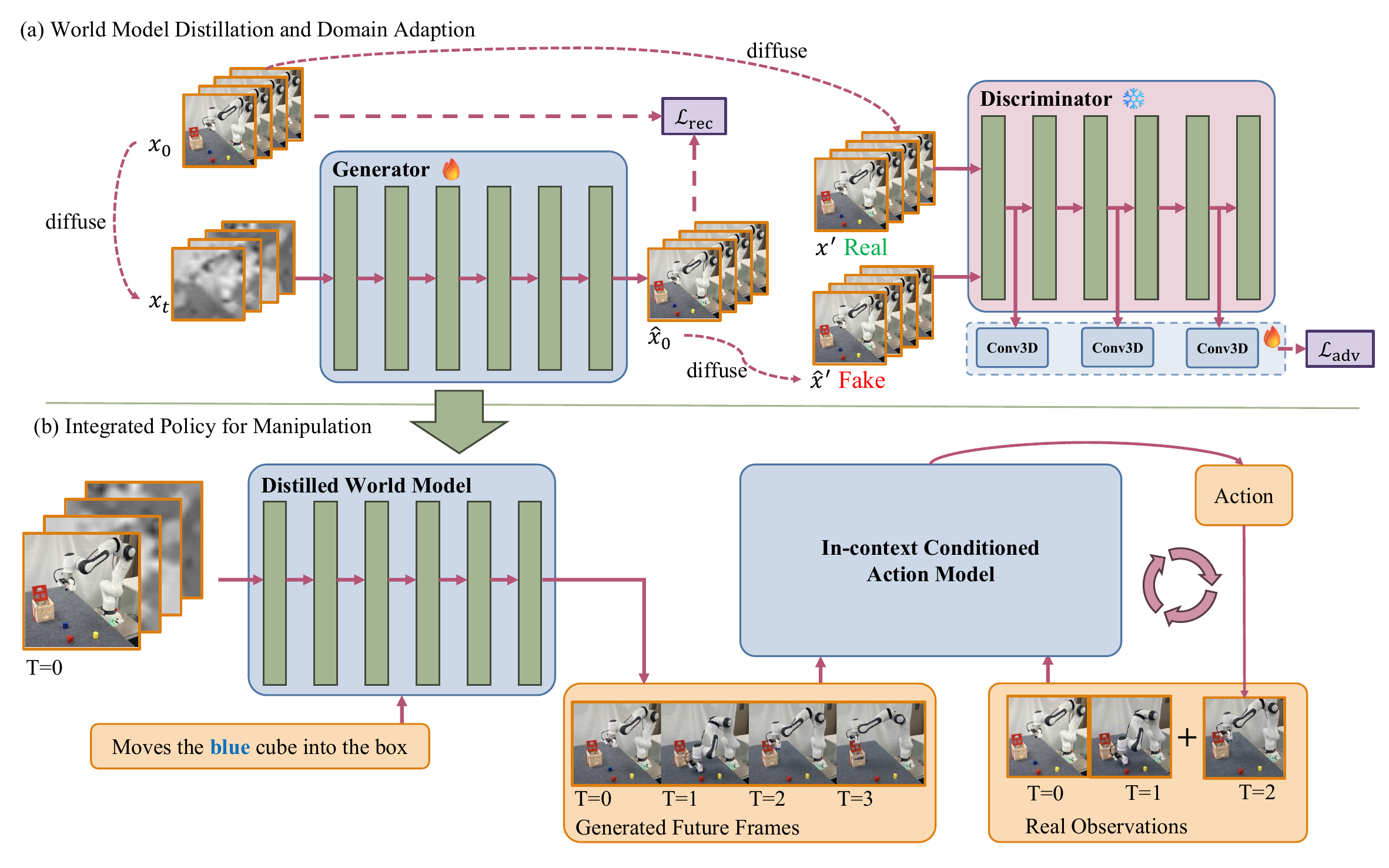}
     \vspace{-0.1in}
   \caption{(a) The training pipeline for world model distillation. Additional details are provided in \cref{sec:distill}. (b) The pipeline of our proposed policy model. Given the current observation and the instruction, the world model first generates imagined future frames. The in-context conditioned action model then produces actions in a closed-loop manner. Each new observation is fed back into the model to generate the next action.   \label{fig:pipeline} }
   \vspace{-0.15in}

\end{figure*}

\section{Related Work}
\label{sec:related_work}

\noindent \textbf{Action Models}. 
Developing action models that can control robots under varied observation conditions has long been a central goal in robotics. Early methods such as ACT~\cite{zhao2023learningfinegrainedbimanualmanipulation} generate action chunks directly from image observations, while Diffusion Policy~\cite{chi2023diffusionpolicy} uses a diffusion-based framework for action generation. More recent approaches, including RT-2~\cite{zitkovich2023rt} and OpenVLA~\cite{kim24openvla}, integrate Vision-Language Models (VLMs) to produce actions directly. However, follow-up work~\cite{black2024pi_0,gr00tn1_2025,wen2025dexvla} shows that VLMs alone often struggle with precise action generation because they are slow at inference and lack a detailed understanding of low-level dynamics. As a result, VLMs are commonly used as high-level decision modules that provide conditions for diffusion- or flow-matching–based action policies. Still, VLMs cannot predict future environmental changes, which motivates combining them with world models that offer predictive visual feedback in addition to high-level decision guidance.

\noindent\textbf{World Models}. 
World models, particularly video or image generation models, have been widely used to enhance action models.  
(1)\textit{World models for fast manipulation.} These models predict future states of the environment to provide visual conditions for action planning. UniPi~\cite{du2023learning} predicts two future frames and applies temporal super-resolution to form short videos for action extraction. Seer~\cite{tian2024predictive} predicts a single future frame along with actions, while GVF-TAPE~\cite{zhang2025generativevisualforesightmeets} additionally predicts depth frames. Unified Video Action Model~\cite{li2025unified} jointly predicts future frames and actions. LaDi-WM~\cite{huang2025ladi} generates geometric and semantic future frames instead of raw images, and FLARE~\cite{zheng2025flare} predicts latent visual features alongside actions. While effective, these models often predict only a few future frames, requiring multiple execution steps to complete tasks and incurring significant time overhead. Video Prediction Policy~\cite{hu2024video} uses one-step denoised latents as visual conditions, but these carry less information than fully denoised frames.
(2)\textit{World models for data generation.} Large video generation models have also been used to generate synthetic training data for pretraining vision-language action models. For example, DreamGen~\cite{jang2025dreamgen} leverages cosmos-predict2 to create large-scale video datasets labeled with neural trajectories. While such models excel at predicting future states, their inference speed is insufficient for `intime' applications. Our Dream4manip runs very fast as shown in results.

\noindent\textbf{Accelerating Diffusion Models}.
Most modern video generation models rely on diffusion pipelines, which require iterative denoising and are computationally expensive. Distillation-based methods have been proposed to accelerate inference. Trajectory-preserving distillation~\cite{salimans2022progressive,liu2022flow,luo2023latent} trains a student model to approximate the teacher’s ODE sampling trajectory with fewer steps, but low-step outputs often suffer from blurriness. Distribution-matching and adversarial distillation address this limitation. Adversarial Diffusion Distillation~\cite{sauer2024adversarial} and LADD~\cite{sauer2024fast} combine adversarial losses with score distillation to improve both speed and quality. SV-F~\cite{zhang2024sf} adapts this approach to video, successfully accelerating SVD~\cite{blattmann2023stable} to one-step generation. Although acceleration may degrade for very large models, adversarial distillation remains a promising strategy for fast, high-quality video prediction.

\section{Method}

Our goal is to enable fast, long-horizon robotic manipulation using a learned world model. As illustrated in \cref{fig:pipeline}, our framework integrates a high-fidelity video-based world model with an action model trained via in-context conditioning. The method consists of three stages: (1) selecting and adapting a robust world model to the robotic domain, (2) distilling the adapted model for efficient few-step denoising, and (3) training an in-context conditioned action model using both imagined and real trajectories. Together, these components form a world-model–driven manipulation policy capable of fast execution.

\subsection{Say: World Model}

\subsubsection{Selection}

\textbf{Key Idea:} Choose a world model that preserves spatial consistency and task-relevant dynamics.  

We benchmark several recent video generation models on metrics critical for embodied decision-making: embodiment consistency, task completion rate, and spatial referring ability. Based on these evaluations, we select \textsc{Cosmos-Predict2} as the backbone of our world model. Quantitative and qualitative comparisons are reported in \cref{sec:gen-compare}. Despite strong performance, the model may degrade in novel scenes, unseen camera viewpoints, or alternative embodiments, motivating domain adaptation and distillation.

\subsubsection{Domain Adaptation and Distillation}
\label{sec:distill}

We introduce a latent-space adversarial loss to enable few-step denoising, reducing computational cost while maintaining prediction fidelity. Even state-of-the-art video generation models can produce spatial inconsistencies when deployed in novel environments. To address this, we first perform domain adaptation to align the world model with the target robotic domain. We then distill the adapted model using a combination of latent adversarial supervision and reconstruction loss, resulting in a world model that is both robust and computationally efficient for fast manipulation.

\noindent \textbf{Denoising Formulation}.
Let $x_0$ denote a 16 dimension encoded video latent. The diffusion transformer $T_\theta$ in \textsc{Cosmos-Predict2} predicts a denoised latent $\hat{x}_0$ from a noisy input $x_t$ conditioned on auxiliary information $cond$:
\begin{equation}
  \hat{x}_0
  =
  c_{skip}\cdot x_t
  +
  c_{out}\cdot
  D_\theta\!\left(
    c_{in}\cdot x^{cond}_t,\; cond
  \right),
  \label{eq:denoising}
\end{equation}
where $x^{cond}_t$ denotes a conditioned version of $x_t$ in which the first
frame is replaced by a clean latent for image-to-video generation. And the preconditioning factors are defined as
\begin{align*}
c_{skip}=(\sigma_{t}+1)^{-1}, & \quad c_{out}=-\frac{\sigma_{t}}{\sigma_{t}+1}\\
c_{in}=c_{skip}, & \quad c_{noise}=-c_{out}
\end{align*}
The noise level $\sigma_t$ is sampled using different strategies during
distillation training and inference.

\noindent \textbf{Noise Schedule for Few-Step Distillation}.
To reduce denoising steps, we sample $\sigma_t$ from a discrete set $\{\sigma_1, \dots, \sigma_{T_g}\}$:
\begin{equation}
  \sigma_t
  =
  \left(
    \sigma_{\min}^{1/p}
    +
    \frac{t}{T_g}
    \left(
      \sigma_{\max}^{1/p}
      -
      \sigma_{\min}^{1/p}
    \right)
  \right)^p,
  \label{eq:gen-sigma}
\end{equation}
where $p$ is a hyper-parameter that controls the emphasis on low noise levels.
The parameter $T_g$ is chosen to match the target number of denoising steps in
the distilled model; unless otherwise specified, we set $T_g = 8$.

Given the sampled noise level $\sigma_t$, we construct a noisy latent
$x_t \sim \mathcal{N}(x_0, \sigma_t)$ and apply \cref{eq:denoising} to obtain the
denoised prediction $\hat{x}_0$.

\begin{figure}
   \includegraphics[width=1.05\linewidth]{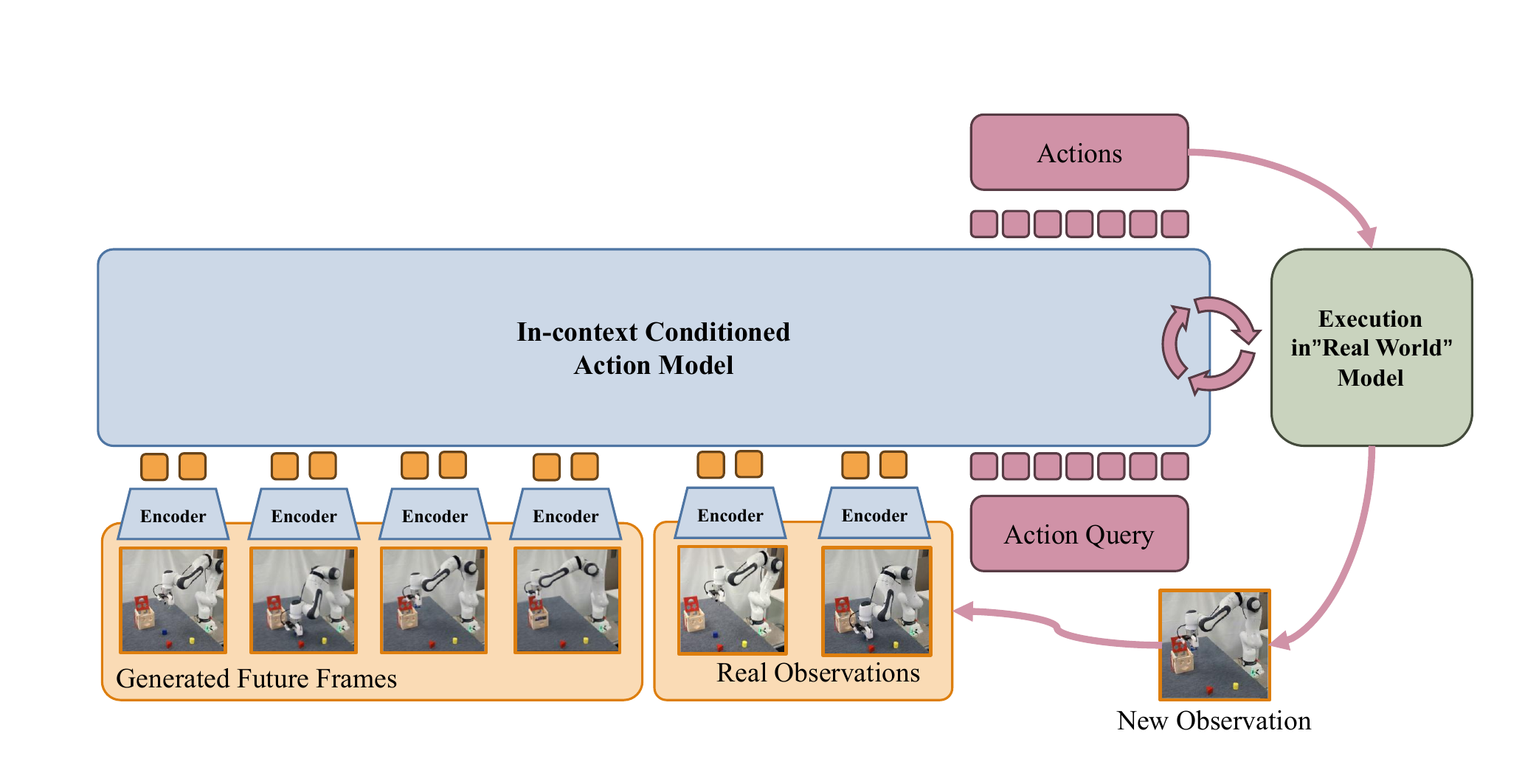}
    \vspace{-0.1in}
   \caption{The structure of the in-context conditioned action model. We use a transformer-based backbone inherited from ACT \cite{zhao2023learningfinegrainedbimanualmanipulation} for our action model, separated vision encoder is assembled to process videos and observations. The model will output an action chunk for each observation.  \label{fig:action-model} }
  
   \vspace{-0.15in}
\end{figure}

\noindent \textbf{Latent Adversarial Distillation}.
To accelerate the iterative denoising process, we introduce a
\emph{latent adversarial loss} that encourages high-fidelity prediction
under a small number of denoising steps.
As illustrated in \cref{fig:pipeline}, we initialize a discriminator
$\mathcal{D}$ using the pretrained diffusion transformer (DiT) weights from
\textsc{Cosmos-Predict2}, and augment it with several lightweight
3D convolutional heads that operate on intermediate DiT feature maps and
produce a pixel-wise score map.
During training, we freeze the DiT backbone and optimize only the added
convolutional heads, encouraging the discriminator to output scores close to
$1$ for clean latents $x_0$ and close to $-1$ for denoised predictions
$\hat{x}_0$.

Before passing $x_0$ or $\hat{x}_0$ to the discriminator, we inject
additional noise by sampling $\sigma^\prime$ such that
$\log(\sigma^\prime) \sim \mathcal{N}(0, 1)$, and diffusing both latents to
obtain $x^\prime$ and $\hat{x}^\prime$, respectively.
This design choice is motivated by the fact that the pretrained DiT is
optimized to process noisy inputs and exhibits more stable and informative
features under moderate noise than on clean latents.

The adversarial objectives for the discriminator $\mathcal{D}$ and the
generator $\mathcal{G}$ are defined as
\begin{equation}
\begin{aligned}
  \mathcal{L}_{adv}^{\mathcal{D}}
  &=
  \mathbb{E}_{x_0,\, \sigma_t,\, \sigma^\prime}
  \Big[
    \mathrm{ReLU}\big(1 - \mathcal{D}(x^\prime)\big)
    +
    \mathrm{ReLU}\big(1 + \mathcal{D}(\hat{x}^\prime)\big)
  \Big], \\
  \mathcal{L}_{adv}^{\mathcal{G}}
  &=
  \mathbb{E}_{x_0,\, \sigma_t,\, \sigma^\prime}
  \big[
    - \mathcal{D}(\hat{x}^\prime)
  \big].
\end{aligned}
\end{equation}

\noindent \textbf{Reconstruction Loss and Domain Adaptation}.
We retain the reconstruction loss $\mathcal{L}_{rec}$ from the
original training objective of \textsc{Cosmos-Predict2} to facilitate
domain adaptation:
\begin{equation}
  \mathcal{L}_{rec}
  =
  \lVert \hat{x}_0 - x_0 \rVert_2^2
  \cdot
  \frac{(1 + \sigma_t)^2}{\sigma_t^2},
  \label{eq:recon}
\end{equation}
where the weighting factor amplifies the contribution of noisier samples,
encouraging accurate denoising across a wide range of noise levels.

In summary, the generator $\mathcal{L}_{\mathcal{G}}$ is distilled by optimizing the combined objective
\[
 \mathcal{L}_{\mathcal{G}}= \lambda\cdot \mathcal{L}_{adv}^{\mathcal{G}} + \mathcal{L}_{rec},
\]
where $\lambda$ is a scaling hyperparameter that balances adversarial
supervision and reconstruction fidelity; unless otherwise stated, we set
$\lambda = 0.1$.

Within the same training iteration, we additionally apply the reconstruction
loss to perform domain adaptation.
Specifically, we sample a noise level $\sigma$ such that
$\log(\sigma) \sim \mathcal{N}(0, 1)$, diffuse the clean latent $x_0$
accordingly, and compute the reconstruction loss between the original latent
$x_0$ and the corresponding denoised prediction $\hat{x}_0$.
This auxiliary objective further aligns the world model with the target
robotic domain while preserving the generative prior.

\begin{table*}
\resizebox{1\linewidth}{!}{
\begin{centering}
\begin{tabular}{cc}
{\small{}}%
\begin{tabular}{c}
{\small{}\hspace{-0.3in}}%
\begin{tabular}{lcccc}
\hline 
\textbf{\small{}Task} & \textbf{\small{}FVD}{\small{} $\downarrow$} & \textbf{\small{}SSIM}{\small{} $\uparrow$} & \textbf{\small{}PSNR}{\small{} $\uparrow$} & \textbf{\small{}LPIPS}{\small{} $\downarrow$}\tabularnewline
\hline 
{\small{}L-10} & {\small{}104.00} & {\small{}0.82 $\pm$ 0.10} & {\small{}21.44 $\pm$ 6.97} & {\small{}0.06 $\pm$ 0.05}\tabularnewline
{\small{}L-Object} & {\small{}177.04} & {\small{}0.89 $\pm$ 0.05} & {\small{}25.30 $\pm$ 5.47} & {\small{}0.04 $\pm$ 0.03}\tabularnewline
{\small{}L-Spatial} & {\small{}119.32} & {\small{}0.85 $\pm$ 0.09} & {\small{}25.16 $\pm$ 6.14} & {\small{}0.03 $\pm$ 0.03}\tabularnewline
{\small{}L-Goal} & {\small{}127.86} & {\small{}0.88 $\pm$ 0.07} & {\small{}26.12 $\pm$ 6.06} & {\small{}0.03 $\pm$ 0.02}\tabularnewline
\hline 
\end{tabular}\tabularnewline
\end{tabular} & {\small{}}%
\begin{tabular}{c}
{\small{}\hspace{-0.35in}}%
\begin{tabular}{llccc}
\hline 
\textbf{\small{}Model} & \textbf{\small{}FVD}{\small{} $\downarrow$} & \textbf{\small{}SSIM}{\small{} $\uparrow$} & \textbf{\small{}PSNR}{\small{} $\uparrow$} & \textbf{\small{}LPIPS}{\small{} $\downarrow$}\tabularnewline
\hline 
{\small{}WAN2.2-14B} & {\small{}714.24} & {\small{}0.63 $\pm$ 0.15} & {\small{}21.49 $\pm$ 3.98} & {\small{}0.11 $\pm$ 0.05}\tabularnewline
{\small{}WAN2.2-14B-Dis} & {\small{}578.62} & {\small{}0.62 $\pm$ 0.18} & {\small{}21.58 $\pm$ 4.46} & {\small{}0.09 $\pm$ 0.06}\tabularnewline
{\small{}C-2B} & {\small{}571.46} & {\small{}0.70 $\pm$ 0.10} & {\small{}22.09 $\pm$ 5.78} & {\small{}0.11 $\pm$ 0.05}\tabularnewline
{\small{}C-14B} & {\small{}522.66} & {\small{}0.67 $\pm$ 0.11} & {\small{}21.56 $\pm$ 5.95} & {\small{}0.12 $\pm$ 0.05}\tabularnewline
{\small{}C-14B-Droid-F.T.} & {\small{}1098.24} & {\small{}0.23 $\pm$ 0.04} & {\small{}14.73 $\pm$ 0.67} & {\small{}0.26 $\pm$ 0.05}\tabularnewline
{\small{}C-2B-DA} & \textbf{\small{}211.84} & {\small{}0.82 $\pm$ 0.09} & {\small{}25.19 $\pm$ 5.48} & {\small{}0.05 $\pm$ 0.03}\tabularnewline
{\small{}C-2B-DA+Dis} & {\small{}238.09} & \textbf{\small{}0.84}{\small{} $\pm$ 0.11} & \textbf{\small{}26.82}{\small{} $\pm$ 6.18} & \textbf{\small{}0.04}{\small{} $\pm$ 0.04}\tabularnewline
\hline 
\end{tabular}\tabularnewline
\end{tabular}\tabularnewline
{\small{}(a) } & {\small{}(b) }\tabularnewline
\end{tabular}
\par\end{centering}
}
\caption{Quantitative results are compared in (a) {\small{}World model evaluation
results of Cosmos-2B on LIBERO benchmarks.} and (b) {\small{}Model
comparison with base model simplified and configuration column (values
rounded to two decimals)}. 'L', and 'C' are short for {\small{}LIBERO}
and Cosmos, DA: Domain Adaptation. \label{tab:fvd} }

\end{table*}

\subsection{Dream: Length-Agnostic World Imagination}
\textbf{Key Idea:} Compress trajectories and generate videos independent of execution length.

We introduce a \emph{length-agnostic imagination} mechanism that compresses arbitrary-length trajectories into a fixed set of keyframes, enabling holistic, frame-rate-agnostic video generation. Unlike prior approaches~\cite{du2023learning,yang2023learning,zhou2024robodreamer} that strictly align video frames with low-level action steps, our insight is that robotic manipulation is fundamentally a low-frequency prediction problem: generating a frame for every action step is unnecessary, redundant, and computationally expensive.

To address this, we compress robotic execution trajectories of arbitrary length into a fixed number of keyframes via uniform temporal sampling:
\begin{equation}
  \mathcal{S}(\tau) = \{x_{t_i}\}_{i=1}^{n}, \quad
  t_i = \left\lfloor \frac{i}{n} \cdot T \right\rfloor,
\end{equation}
where $n \ll T$ is a predefined constant independent of the trajectory length $T$. In our experiments, we set $n = 93$.

We then finetune a video diffusion model to learn a frame-rate-agnostic video generation capability, enabling holistic imagination of the entire execution process without dependence on the specific execution length. This allows the model to anticipate long-horizon outcomes while remaining computationally efficient.

\subsection{Act: In-Context Conditioned Action Model}

\textbf{Key Idea:} Treat imagined trajectories as in-context examples to guide action prediction.

Even with state-of-the-art video generation models, the distilled world model may exhibit occasional spatial inaccuracies. To address this, we treat imagined trajectories as \emph{in-context examples} for action prediction rather than strict commands~\cite{vuong2025action}, allowing the action model to ground its predictions in real observations and correct errors in the generated video.

As illustrated in \cref{fig:action-model}, our action model receives both the generated future frames and the real historical observations. This design allows the model to correct spatial errors present in the generated video by continuously grounding its predictions in actual observations. Functionally, the model operates similarly to a next-token predictor: given an execution example in the form of generated frames, it outputs actions whose execution produces real-world observations that follow the general pattern demonstrated in the example.

\begin{table}[htbp] \small
\centering

\begin{tabular}{lcccc}
\toprule
\textbf{Model} & 
\textbf{EC $\uparrow$} &
\textbf{RSR $\uparrow$ } & 
\textbf{ISR $\uparrow$ } & 
\textbf{TSR $\uparrow$ } \\
\midrule
Cosmos-2B   & 1.58 & \textbf{96.00} & \textbf{86.00} & 70.00 \\
Cosmos-14B  & \textbf{1.62} & 90.00 & 82.00 & \textbf{76.00} \\
Cosmos-14B-Droid   & 1.58 & 78.00 & 58.00 & 34.00 \\
Wan-14B            & 1.56 & 54.00 & 44.00 & 20.00 \\
\bottomrule
\end{tabular}
\caption{Percentage-based evaluation across Image-to-Video models, with best scores in bold.\label{tab:percent_metrics}}
\vspace{-0.2in}
\end{table}

\section{Experiments}

\begin{table*} \small
\caption{Quantitative comparison on the LIBERO~\cite{liu2023libero} benchmark. Param denotes the scale of the VLM backbone.  \label{tab:libero}
}
\centering
\resizebox{0.7\textwidth}{!}{
\begin{tabular}{l|c|cccc|c} 
\hline
\multirow{2}{*}{Method} & 
\multirow{2}{*}{Param} & \multicolumn{4}{c|}{Task Suite} & \multirow{2}{*}{Total} \\
 & & Spatial & Object & Goal & Long & \\
\hline
OpenVLA~\cite{kim2025openvla} & 7 & 84.7 & 88.4 & 79.2 & 53.7 & 76.5 \\
OpenVLA-OFT~\cite{kim2025fine} & 7 & 97.6 & 98.4 & 97.9 & 94.5 & 97.1 \\
CoT-VLA~\cite{zhao2025cot} & 7 & 87.5 & 91.6 & 87.6 & 69.0 & 81.1 \\
UniVLA~\cite{bu2025univla} & 7 & 96.5 & 96.8 & 95.6 & 92.0 & 95.2 \\
WorldVLA~\cite{cen2025worldvla} & 7 & 87.6 & 85.2 & 75.1 & 54.1 & 74.8 \\
4D-VLA~\cite{zhang20254d} & 4 & 88.9 & 95.2 & 90.9 & 79.1 & 88.6 \\
SpatialVLA~\cite{qu2025spatialvla} & 4 & 88.2 & 95.2 & 90.9 & 79.1 & 88.6 \\
$\pi_0$~\cite{black2024pi0} & 3 & 96.8 & 98.8 & 95.8 & 85.2 & 94.2 \\
$\pi_0$-FAST~\cite{pertsch2025fast} & 3 & 96.4 & 96.8 & 88.6 & 60.2 & 85.5 \\
SmolVLA~\cite{shukor2025smolvla} & 2.25 & 93.0 & 94.0 & 91.0 & 77.0 & 88.8 \\
GR00t N1~\cite{bjorck2025gr00t} & 2 & 94.4 & 97.6 & 93.0 & 90.6 & 93.9 \\
DreamVLA~\cite{zhang2025dreamvla} & 0.57 & 97.5 & 94.0 & 89.5 & 89.5 & 92.6 \\
\hline
\textbf{Dream4manip (Ours)} & 1.22 & 99.4 & 99.2 & 98.2 & 95.4 & 98.1 \\
\hline
\end{tabular}
}
\end{table*}

\textbf{Experimental Setup}. We evaluate our approach in two stages. First, we assess the world model by testing the zero-shot spatial referring capabilities of four baseline video generation models—Cosmos-2B, Cosmos-14B, Cosmos-Droid, and Wan-14B—using a task-specific metric to select the best candidate for domain adaptation. We further validate improvements through real-world qualitative analyses after adaptation and adversarial distillation. Second, we perform rollouts in both simulated and real-world tasks to evaluate the effectiveness of our proposed policy in guiding robust and precise manipulation.

\mysubsub{Simulation benchmark}
In the assessment of the whole policy, we conduct close loop evaluation on LIBERO benchmark \cite{liu2023libero}.
For the LIBERO benchmark, each task suite contains approximately 500 training samples (Specifically: Spatial—433, Object—456, Goal—436, and Long—389) and 10 test subtask samples. During evaluation, each test sample is executed 50 times, resulting in a total of 500 test runs per suite.

\begin{figure*}
  \centering
   \includegraphics[width=0.8\linewidth]{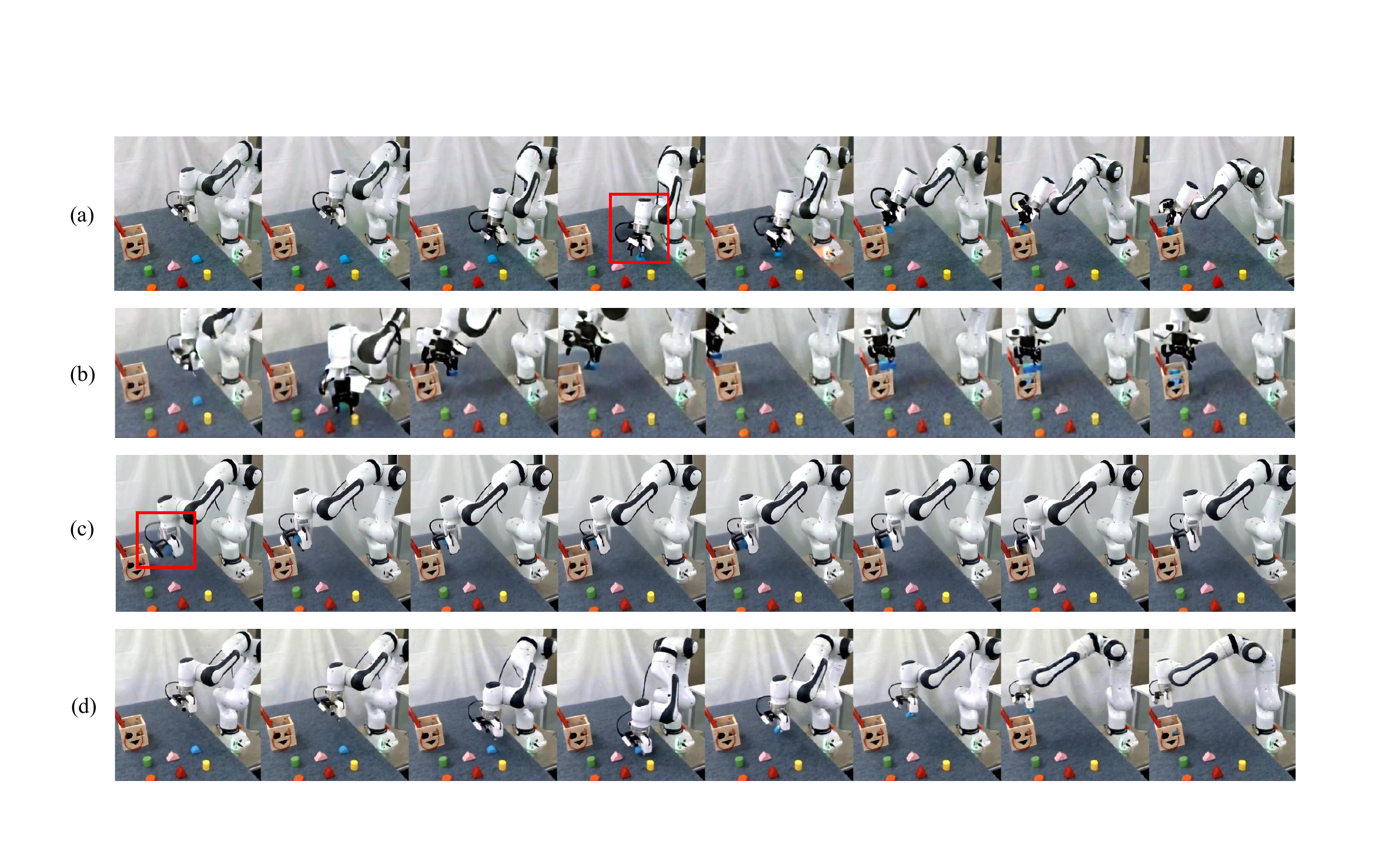}
   \vspace{-0.3in}
   \caption{\small Qualitative results generated by different video generation models for the instruction “Move the blue block into the wooden box.”
(a) Output from the Cosmos-predict2 2B model. The model incorrectly replaces the robot gripper with a Robotiq-style two-finger gripper, likely reflecting common embodiments in its training data.
(b) Output from the Cosmos-predict2 14B model fine-tuned on DROID. This version predicts multiview videos from left, right, and wrist cameras, but the camera poses still differ from our test setup, leading to incorrect arm scale.
(c) Output from the Wan2.2 14B model. Wan struggles to interpret complex instructions: it fails to infer that the robot should first pick up the block before moving it, and instead assumes the robot is already holding the blue block while moving toward the box.
(d) Output from the Cosmos-predict2 2B model after domain adaptation. Domain adaptation enables the model to generate future predictions that remain consistent with the robot’s true embodiment.   \label{fig:flaws-WM}  }
 \vspace{-0.15in}
\end{figure*}

\mysubsub{Real-world setup}
To evaluate the effectiveness of our method in real-world settings, we build a physical robotic experimental setup as illustrated in \cref{fig:flaws-WM}. The platform employs a Franka 7-DoF robotic arm as the manipulation agent. An Intel RealSense D435 camera is mounted in a third-person view to provide visual observations of the robot. Teleoperation data are collected using a 3Dconnexion SpaceMouse.

\noindent \textbf{Implementation Details}.
For the world model, we train our model based on cosmos-predict2. We adopt FusedAdamW as the optimizer with a learning rate of $2^{-14.5}$. A lambda linear scheduler is used, with zero warm-up steps, a single-cycle length of 1{,}000 steps, $f_{\max}=0.6$, and $f_{\min}=0$. Training is conducted with FSDP-based distributed parallelism, where the context parallel size is set to 2, and exponential moving average (EMA) is enabled to stabilize convergence. The world model is trained for 10{,}000 iterations on the LIBERO benchmark, while real-world training is performed for 1{,}000 iterations.  When training the action model, we initialize the VLM with Qwen2.5 and perform parameter-efficient fine-tuning of the VLM backbone using LoRA with a rank of 64. Training is conducted on 8 GPUs with a total batch size of 128. During training, we incorporate future video predictions generated by the world model as additional contextual inputs, and uniformly sample 8 frames from historical video segments as conditioning context. The learning rate is set to $2 \times 10^{-4}$, and the total number of training steps is 20{,}000.

\begin{figure}[htbp]
  \centering
   \includegraphics[width=1\linewidth]{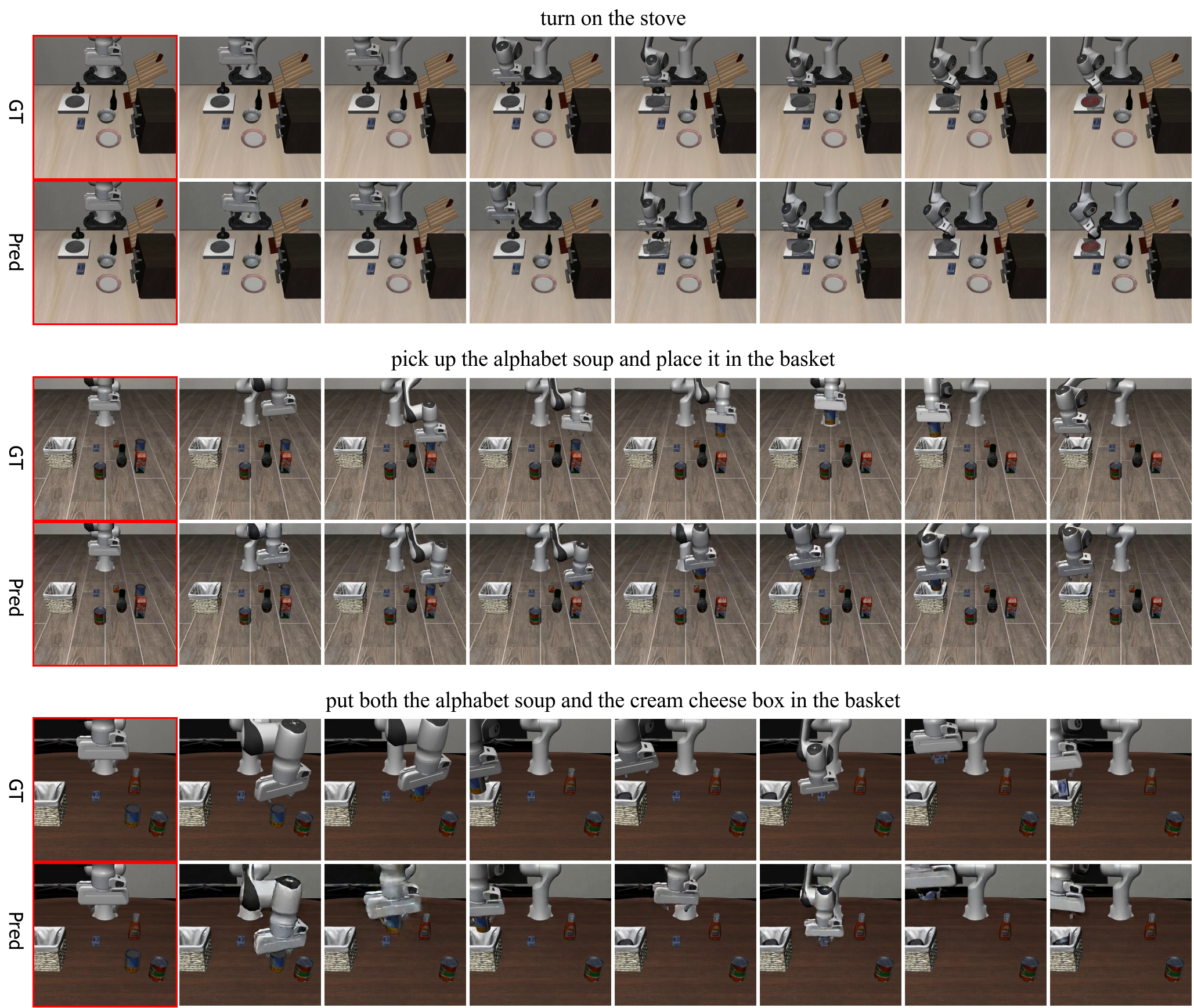}
    \vspace{-0.05in}
   \caption{\small \textbf{Qualitative results on LIBERO}. We show world model generation results across different tasks. The red-bordered frame denotes the conditioning frame, while the remaining frames are generated.  \label{fig:quat_libero} }

    \vspace{-0.15in}
\end{figure}

\begin{table}[htbp]
\begin{centering}
\resizebox{\linewidth}{!}{
\begin{tabular}{cccccc}
\hline 
\textbf{\small{}D} & \textbf{\small{}Time(s)} & \textbf{\small{}FVD}{\small{} $\downarrow$} & \textbf{\small{}SSIM}{\small{} $\uparrow$} & \textbf{\small{}PSNR}{\small{} $\uparrow$} & \textbf{\small{}LPIPS}{\small{} $\downarrow$}\tabularnewline
\hline 
{\small{}1} & {\small{}10.72} & {\small{}1128.10} & {\small{}0.54 $\pm$ 0.14} & {\small{}17.89 $\pm$ 6.95} & {\small{}0.20 $\pm$ 0.07}\tabularnewline
{\small{}5} & {\small{}19.43} & {\small{}1445.19} & {\small{}-0.02 $\pm$ 0.29} & {\small{}12.43 $\pm$ 8.43} & {\small{}0.30 $\pm$ 0.10}\tabularnewline
{\small{}10} & {\small{}29.80} & {\small{}579.90} & {\small{}0.70 $\pm$ 0.09} & {\small{}22.10 $\pm$ 5.78} & {\small{}0.11 $\pm$ 0.05}\tabularnewline
{\small{}20} & {\small{}52.59} & {\small{}496.06} & \textbf{\small{}0.72 $\pm$ 0.09} & \textbf{\small{}22.48 $\pm$ 5.67} & \textbf{\small{}0.10 $\pm$ 0.04}\tabularnewline
{\small{}35} & {\small{}86.20} & \textbf{\small{}488.06} & \textbf{\small{}0.72 $\pm$ 0.09} & {\small{}22.39 $\pm$ 5.70} & \textbf{\small{}0.10 $\pm$ 0.04}\tabularnewline
\hline 
\end{tabular}
}
\caption{Ablation study on different denoise steps (D) on Cosmos-predict2 2B
model. Metrics are rounded to two decimals.}

\label{tab:diff-step}
\par\end{centering}
\end{table}

\subsection{Zero-shot Spatial Referring Abilities.}

We evaluate four base image-to-video generation models on their spatial referring ability using four metrics on the task “Pick the block of a specific color and place it.”

\textbf{Embodiment Consistency (EC)} is quantified by analyzing the frequency of three predefined cases in the generated videos, reflecting different levels of embodiment stability over time. Specifically, \textbf{Case I} corresponds to scenarios where the embodiment remains fully consistent across all frames and is assigned 3 points; \textbf{Case II} captures situations in which the embodiment undergoes changes but remains recognizable, receiving 2 points; and \textbf{Case III} denotes cases where the embodiment becomes difficult to recognize or is heavily distorted, which are assigned 1 point.

EC is the average score across all generated videos. A higher EC indicates better consistency in predicting the robot embodiment.

\textbf{Referring Success Rate (RSR)} measures whether the generated embodiment moves toward the correct target object based on its color. A success is recorded when the embodiment moves more than halfway toward the correct object.

\textbf{Interaction Success Rate (ISR)} evaluates whether the embodiment successfully grasps the correct target object during the interaction. An interaction is considered a failure if any of the following conditions occur: (1) the target object moves unnaturally toward the gripper without physical contact; (2) the gripper fails to reach the target object’s position relative to the first frame; or (3) the gripper becomes severely distorted, rendering the interaction outcome ambiguous.

\textbf{Task Completion Rate (TCR)} counts a success when the generated video shows the target object being correctly placed into the wooden box.

We demonstrate those evaluation metrics in \cref{tab:percent_metrics}. Overall, the original Cosmos2 series models exhibit stronger spatial referring and manipulation abilities. In contrast, Cosmos-14B-Droid suffers from a domain gap due to multiview fine-tuning, and the Wan model performs poorly because it lacks manipulation-focused training.

The qualitative results can be observed in \cref{fig:flaws-WM}. 
General-purpose video generation models such as WAN often fail to faithfully capture robot embodiment and task semantics, leading to incorrect gripper structures, mismatched camera poses, or flawed action reasoning. 
Large-scale pretrained Cosmos models exhibit better physical motion modeling, but still struggle to align with the real execution setup, whereas domain-adapted models produce predictions that remain consistent with the true robot configuration and task dynamics.

\subsection{Generation Quality  among Video Models \label{sec:gen-compare} }

We evaluate the generation quality of several video models, with quantitative results reported in Table~\ref{tab:fvd}. Across all metrics, the raw Cosmos models outperform the Wan model. Applying domain adaptation yields substantial improvements: for instance, FVD drops from 571.46 (C-2B) to 211.84 (C-2B-DA), while SSIM rises from 0.70 to 0.82, reflecting enhanced temporal coherence and frame-level fidelity. Notably, Cosmos-14B-Droid-F.T., despite being fine-tuned on robotic data, performs worse than the base Cosmos models (FVD 1098.24, SSIM 0.23). We attribute this degradation to the domain gap introduced by its multi-view training, which assumes camera configurations differing from our single third-person observations. This highlights that naive fine-tuning on robotic data can hurt performance when observation setups mismatch, motivating our targeted domain adaptation approach.

Table~\ref{tab:fvd}(a) further shows consistent performance of our adapted world model across the LIBERO task suites. L-Goal achieves the highest PSNR (26.12), while L-Object exhibits the strongest SSIM (0.89) and lowest FVD (177.04). The relatively uniform metrics across all suites suggest that our adapted model generalizes effectively to manipulation tasks of varying complexity. Combining domain adaptation with latent adversarial distillation (C-2B-DA+Dis) in Table~\ref{tab:fvd}(b) maintains strong performance while enabling efficient few-step inference: SSIM improves slightly from 0.82 to 0.84, PSNR increases from 25.19 to 26.82, and FVD experiences a minor increase (211.84 $\rightarrow$ 238.09), representing a modest trade-off for significant computational gains.

\subsection{Close-Loop Policy Evaluation}

We first present the world model generation results in \cref{fig:quat_libero}. As shown, our model successfully predicts the task execution process by generating complete sequences of robot manipulation frames, demonstrating its ability to anticipate future trajectories. In particular, the third task in \cref{fig:quat_libero}, which requires picking up the alphabet soup can before placing the cream cheese box, represents a long-horizon manipulation challenge that our model handles effectively. Qualitative results in \cref{tab:fvd} further confirm this, with low FVD scores across all four task suites.

Table~\ref{tab:libero} reports quantitative results across the LIBERO task suites, evaluating different types of generalization. All the methods are trained and evaluated under the same settings. Dream4manip consistently achieves the best overall performance, reaching a total success rate of 98.2\% while using a moderate-sized backbone compared to large-scale VLA models.

On LIBERO-Spatial, Dream4manip achieves a success rate of 99.4\%, demonstrating strong robustness to spatial variations. 
On LIBERO-Object, it maintains 99.2\%, indicating effective object-level generalization. 
For LIBERO-Goal, it reaches 98.6\%, highlighting the benefit of imagination for multi-stage reasoning. 
On the challenging LIBERO-Long suite, it achieves 95.4\%, confirming its strength on long-horizon tasks. 
Overall, these results show that a video-based world model significantly improves closed-loop control across diverse generalization settings.

\subsection{Ablation of denoising steps}

We analyze the effect of different denoising steps when generating videos using the Cosmos2 2B model, which is originally trained with 35 denoising steps for high-fidelity video synthesis. As shown in \cref{tab:diff-step}, SSIM and PSNR increase sharply, while FVD and LPIPS decrease sharply as the number of denoising steps increases from 1 to 10. Beyond 10 steps, the improvement slows and the metrics fluctuate around a stable value. Overall, this reveals a trade-off between video quality and generation speed, with 10 denoising steps offering a good balance.

\textbf{Discussion}. While our results demonstrate significant improvements, several challenges remain. First, although adversarial distillation accelerates video prediction, achieving low-latency, fast control on resource-constrained robots requires further optimization. Second, long-horizon predictions can still accumulate errors, particularly in dynamic or cluttered environments; online correction, hybrid planning, or uncertainty-aware rollouts may help mitigate drift. Third, the performance of our in-context conditioned action model depends on the fidelity of generated videos, motivating methods robust to imperfect or partially inconsistent rollouts. 

\textbf{Future Work}. Looking forward, scaling to diverse embodiments and complex multi-step tasks remains a key challenge. Generalization across radically different robots, viewpoints, or environments may require broader pretraining and stronger embodiment-conditioned representations. Moreover, integrating predictive world models with language-driven policies offers a path toward compositional, instruction-aware manipulation, enabling robots to anticipate, plan, and adapt in novel scenarios—an important step toward general-purpose, adaptive embodied intelligence.

\section{Conclusion}
We presented a unified framework that combines accelerated world modeling with in-context conditioned action prediction to enable precise, efficient robotic manipulation. Our three-stage design—world model selection and adaptation, adversarial distillation for fast video prediction, and in-context action modeling—overcomes key limitations of prior approaches, including limited temporal foresight, spatial inconsistency, and high computational cost. Extensive evaluation shows that high-quality, long-horizon video predictions significantly improve downstream action generation, with Cosmos2 models benefiting from domain adaptation and adversarial distillation to yield more reliable and robust manipulation. By bridging predictive visual foresight and control, our framework moves toward adaptive, anticipatory robots capable of planning and acting effectively in complex, real-world environments.

\section*{Impact Statement}

We enable robots to anticipate and act on future states by combining fast, predictive video generation with action modeling, significantly improving spatial reasoning, embodiment consistency, and task performance in real-world manipulation. While our work could influence a range of societal applications from manufacturing automation to assistive robotics, we do not identify any immediate ethical or safety concerns that require special emphasis here.

\nocite{langley00}

\bibliography{main}
\bibliographystyle{icml2026}

\newpage
\appendix
\onecolumn
\section{Detailed definition of spatial referring abilities metrics.}
We evaluate four base image-to-video generation models on their spatial referring ability using four metrics:
Embodiment Consistency (EC), Referring Success Rate (RSR), Interaction Success Rate (ISR) and Task Completion Rate (TCR). Here we give the accurate definition and examples for these metrics.

\textbf{Embodiment Consistency (EC)} is quantified by analyzing the frequency of three predefined cases in the generated videos, reflecting different levels of embodiment stability over time. Specifically, \textbf{Case I} corresponds to scenarios where the embodiment remains fully consistent across all frames and is assigned 3 points; \textbf{Case II} captures situations in which the embodiment undergoes changes but remains recognizable, receiving 2 points; and \textbf{Case III} denotes cases where the embodiment becomes difficult to recognize or is heavily distorted, which are assigned 1 point.

EC is the average score across all generated videos. A higher EC indicates better consistency in predicting the robot embodiment.

\textbf{Referring Success Rate (RSR)} measures whether the generated embodiment moves toward the correct target object based on its color. A success is recorded when the embodiment moves more than halfway toward the correct object.

\textbf{Interaction Success Rate (ISR)} evaluates whether the embodiment successfully grasps the correct target object during the interaction. An interaction is considered a failure if any of the following conditions occur: (1) the target object moves unnaturally toward the gripper without physical contact; (2) the gripper fails to reach the target object’s position relative to the first frame; or (3) the gripper becomes severely distorted, rendering the interaction outcome ambiguous.

\textbf{Task Completion Rate (TCR)} counts a success when the generated video shows the target object being correctly placed into the wooden box.

\section{Out-of-distribution test on spatial referring abilities of the selected world model.}
\begin{figure*}
  \centering
   \includegraphics[width=0.9\linewidth]{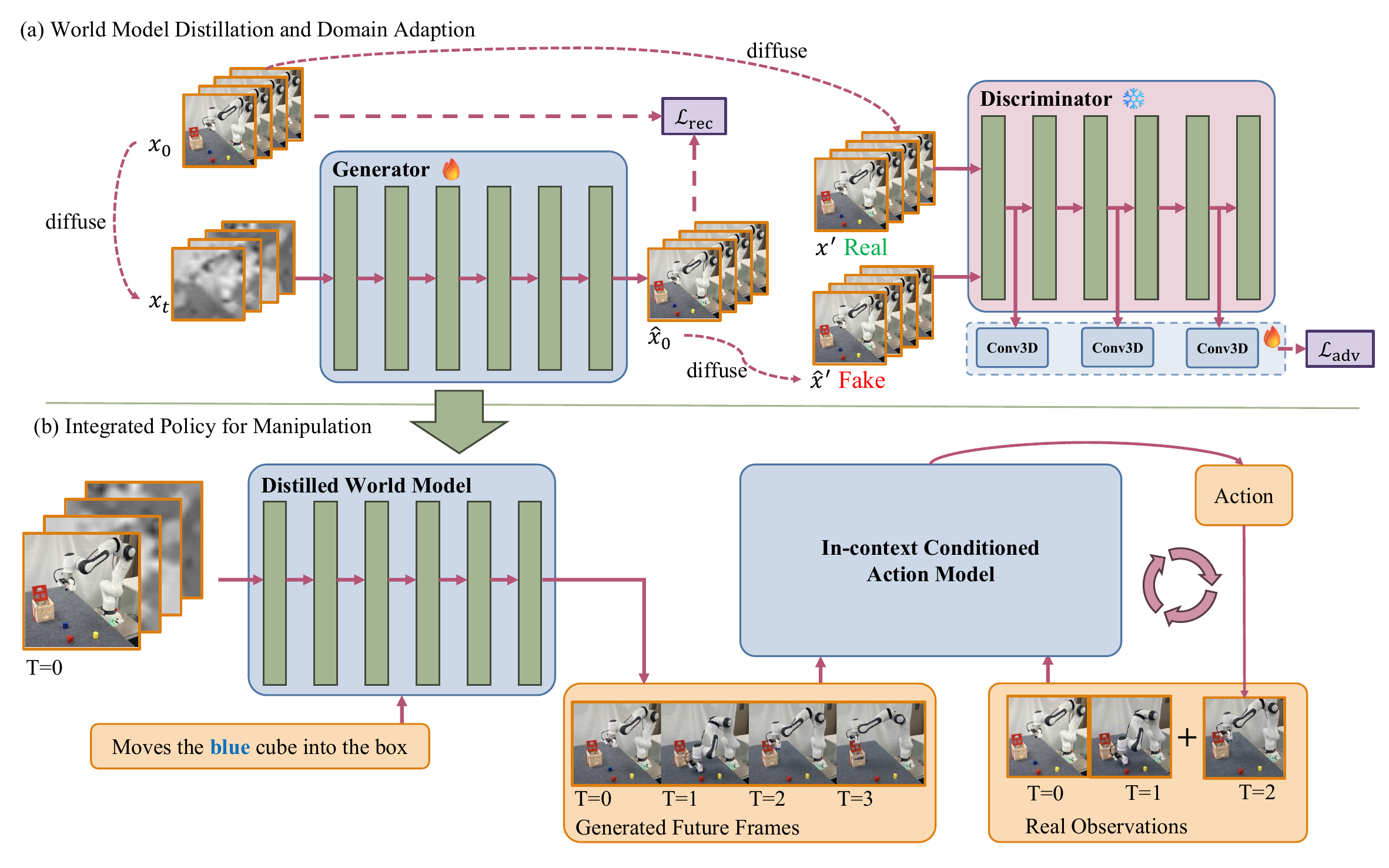}
     \vspace{-0.1in}
   \caption{ \label{fig:RBY} }
   \vspace{-0.15in}

\end{figure*}

As shown in \cref{fig:RBY}, we place 3 blocks with different colors (red, blue and yellow) 

\end{document}